# Ontology-Based Knowledge Modeling and Uncertainty-Aware Outdoor Air Quality Assessment Using Weighted Interval Type-2 Fuzzy Logic

Md Inzmam, Ritesh Chandra, Sadhana Tiwari, Sonali Agarwal, Triloki Pant

*Department of Information Technology, Indian Institute of Information Technology Allahabad Prayagraj, India*

**Abstract**

Outdoor air pollution is a major concern for the environment and public health, especially in areas where urbanization is taking place rapidly. The Indian Air Quality Index (IND-AQI), developed by the Central Pollution Control Board (CPCB), is a standardized reporting system for air quality based on pollutants such as $PM_{2.5}$, $PM_{10}$), nitrogen dioxide ($NO_2$), sulfur dioxide ($SO_2$), ozone ($O_3$), carbon monoxide (CO), and ammonia ($NH_3$). However, the traditional calculation of the AQI uses crisp thresholds and deterministic aggregation rules, which are not suitable for handling uncertainty and transitions between classes. To address these limitations, this study proposes a hybrid ontology-based uncertainty-aware framework integrating Weighted Interval Type-2 Fuzzy Logic with semantic knowledge modeling. Interval Type-2 fuzzy sets are used to model uncertainty near AQI class boundaries, while pollutant importance weights are determined using Interval Type-2 Fuzzy Analytic Hierarchy Process (IT2-FAHP) to reflect their relative health impacts. In addition, an OWL-based air quality ontology extending the Semantic Sensor Network (SSN) ontology is developed to represent pollutants, monitoring stations, AQI categories, regulatory standards, and environmental governance actions. Semantic reasoning is implemented using SWRL rules and validated through SPARQL queries to infer AQI categories, health risks, and recommended mitigation actions. Experimental evaluation using CPCB air quality datasets demonstrates that the proposed framework improves AQI classification reliability and uncertainty handling compared with traditional crisp and Type-1 fuzzy approaches, while enabling explainable semantic reasoning and intelligent decision support for air quality monitoring systems.

## 1. Introduction

Air pollution has emerged as one of the most serious environmental and public health issue globally, especially in rapidly urbanizing nations such as India. Chronic inhalation of high concentrations of ambient air pollutants like particulate matter $PM_{2.5}$, $PM_{10}$, $NO_2$, $SO_2$, $O_3$, CO, and $NH_3$ it found to be strongly associated respiratory diseases, cardiovascular diseases, and early death [1]. To address these rising concerns, effective air quality monitoring and communication tools are essential for improving public awareness and facilitating informed decision-making.

In the Indian context, the CPCB monitors the ambient air quality through a network of air quality monitoring stations across the country. The CPCB measures air quality in terms of the IND-AQI, which combines the concentrations of various pollutants into a single value that falls into linguistic categories such as *Good, Satisfactory, Moderate, Poor, Very Poor,* and *Severe* [2][3]. While this form of representation makes it easier for the common man to understand complex air quality information, the classification procedure is inherently ambiguous, especially when pollutant concentrations lie near category boundaries.

Traditional air quality assessment methods use sharp thresholds and deterministic rules, requiring sharp transitions between AQI classes. These methods cannot properly capture the gradual, uncertain, and imprecise nature of real-world air quality data [4]. Traditional ontology-based semantic systems, which are useful for structured knowledge representation and interoperability, also use binary logic and thus cannot capture the uncertainty and vagueness that naturally exist in environmental phenomena [5][6].

Fuzzy logic has been extensively used to overcome these drawbacks by allowing partial membership of pollutant concentrations in several AQI classes simultaneously [7]. However, Type-1 fuzzy systems require sharp membership functions and fixed strengths of rule activation, which are prone to uncertainty caused by sensor noise, expert vagueness, and spatiotemporal variability of pollutants [8]. In outdoor air quality assessment, where the effects of pollutants vary in terms of severity and health significance, assigning equal weights to all inference rules may result in suboptimal and biased AQI assessment.

To address these challenges, Type-2 fuzzy logic provides a more expressive paradigm that formally addresses the issue of uncertainty modeled in membership functions them-

---
*Email addresses:* prf.minzmam@iiita.ac.in (Md Inzmam), rsi2022001@iiita.ac.in (Ritesh Chandra), prf.sadhana@iiita.ac.in (Sadhana Tiwari), sonali@iiita.ac.in (Sonali Agarwal), tpant@iiita.ac.in (Triloki Pant)

selves [9][10]. Additionally, the use of weighted rule firing enables the inference system to consider the relative significance of various pollutants and rules, based on their differing toxicities and effects on overall air quality [11]. The use of weights on fuzzy rules enables the inference system to be more robust, interpretable, and relevant to real-world concerns.

On the other hand, ontologies are a powerful tool for the formal representation of air quality knowledge, pollutant information, AQI values, and their relationships. When combined with fuzzy logic, fuzzy ontologies support semantic reasoning on uncertain and imprecise information, turning raw numerical data into meaningful and human-understandable knowledge [13].

In this paper, we present a weighted interval Type-2 fuzzy ontology-based framework for outdoor air quality evaluation using CPCB air quality monitoring data. The proposed framework combines:

- Interval Type-2 fuzzy inference for dealing with uncertainties in pollutant identification,

- Weighted rule firing for representing the relative importance of air pollutants and inference rules, and

- Semantic ontology modeling for facilitating structured knowledge representation, annotation, and reasoning on outdoor air quality information.

The proposed framework differs from the existing AQI computation approaches in that it semantically enriches CPCB air quality data and enables intelligent reasoning on ambiguous AQI situations. The proposed framework aims to improve the transparency, reasoning, and decision-making capabilities of outdoor air quality monitoring and analysis in the Indian scenario.

The key contributions of this research work are listed below:

- A hybrid weighted interval Type-2 fuzzy inference model complying with CPCB AQI standards for outdoor air quality assessment.

- Incorporation of rule weighting approaches for enhancing the accuracy and interpretability of AQI assessment.

- Design of a Type-2 fuzzy ontology for semantic representation and reasoning of outdoor air quality knowledge.

- Transformation of raw air quality data into RDF triples and development of SWRL rules for rule-based decision support using classes, properties, and individuals.

- Validation of the proposed framework using real-world CPCB air quality data from Indian monitoring stations.

The rest of this paper is organized as follows. Section 2 discusses the related work on fuzzy air quality assessment and semantic modeling. Section 3 explains the proposed weighted Type-2 fuzzy inference approach. Section 4 describes the ontology design and system architecture. Section 5 discusses the experimental results and performance analysis using CPCB data. Finally, Section 6 concludes this paper and identifies future research directions.

## 2. Related Work

Air quality evaluation has been an area of interest because of its close link with human health and sustainability. Various methods have been proposed to deal with the uncertainty involved in air quality measurements. The literature related to the current research work can be classified into three categories: air quality evaluation using fuzzy logic, air quality evaluation using Type-2 fuzzy inference systems with weighted reasoning, and air quality evaluation using ontology-based semantic models.

### 2.1. Fuzzy Logic–Based Air Quality Assessment

Traditional air quality assessment methods usually involve crisp threshold classification, where air pollutant concentrations are assigned to distinct AQI classes. Nevertheless, these crisp classification methods are not capable of handling the transition zones and uncertainties that exist in the vicinity of class boundaries. To overcome this drawback, fuzzy logic was proposed as a useful paradigm for representing linguistic variables and imprecise knowledge.

Based on this background, various research works have utilized Type-1 fuzzy inference systems for air quality assessment. Gorai et al. showed that fuzzy logic-based AQI models are more flexible and robust than traditional index-based models since they enable partial membership in multiple AQI classes [4]. Similarly, Yadav et al. proposed a fuzzy inference system for air quality assessment and concluded that fuzzy-based classification is more appropriate than traditional crisp classification to model real-world air quality conditions that are inherently uncertain and varying [8]. Haque and Singh presented a detailed review of various AQI calculation models and emphasized the superiority of fuzzy-based models over crisp models [3].

However, despite the above benefits, Type-1 fuzzy systems require exact membership functions and equal weights for rules, which hinders their capacity to deal with uncertainties associated with sensor noise, expert subjectivity, and spatio-temporal variability of outdoor air pollutants.

### 2.2. Type-2 Fuzzy Inference and Weighted Rule-Based Approaches

However, the drawback of Type-1 fuzzy logic led to the development of Type-2 fuzzy sets, which can handle uncertainties directly in membership functions. Mendel and John introduced the concept of interval Type-2 fuzzy sets to handle uncertainties in a computationally feasible manner [9]. Mendel showed that Type-2 fuzzy systems perform better in decision-making tasks with ambiguity and noise [10].

In the context of air quality assessment, Debnath et al. proposed a weighted interval Type-2 fuzzy inference system, emphasizing that different air pollutants contribute unequally to overall air quality and associated health risks [11]. Their work showed that incorporating rule weights improves AQI evaluation accuracy and interpretability. Rule-based fuzzy systems improve interpretability and structured decision-making by leveraging expert-defined rules, as demonstrated in intelligent real-time prediction frameworks [12].

The theoretical foundations of weighted and uncertainty-aware fuzzy reasoning have been extensively discussed in the fuzzy systems literature. Mendel introduced rule-based fuzzy systems capable of handling uncertainty in both rules and membership functions [14], while perceptual computing further formalized the role of weighted reasoning in subjective decision-making processes [11]. Efficient computational realization of Type-2 fuzzy inference has been supported by enhanced Karnik–Mendel algorithms [15].

While weighted Type-2 fuzzy inference systems enhance numerical AQI computation, they primarily focus on quantitative assessment and lack semantic representation and reasoning mechanisms for structured knowledge interpretation.

*2.3. Ontology-Based and Semantic Air Quality Modeling*

Ontology-based approaches have been widely adopted for structured knowledge representation and semantic interoperability across heterogeneous data sources. Gruber defined ontology as a formal and explicit specification of a shared conceptualization, forming the theoretical basis for ontology engineering [5]. Ontology languages such as OWL further enable reasoning and interoperability in semantic web applications [6].

In environmental and air quality domains, ontologies have been used to represent pollutants, AQI categories, monitoring stations, and health impacts. However, classical ontologies rely on binary logic and are inherently incapable of modeling uncertainty and vagueness present in real-world environmental data.

To address this limitation, fuzzy ontologies have been proposed by integrating fuzzy logic with semantic web technologies. Bobillo and Straccia introduced fuzzy ontology representations using OWL 2, enabling uncertainty-aware semantic reasoning [16], while Straccia further formalized the foundations of fuzzy logic in semantic web languages [17]. More recently, Type-2 fuzzy ontology–based frameworks have demonstrated improved capability in handling higher-order uncertainty in air quality assessment tasks by combining semantic modeling with Type-2 fuzzy reasoning [13].Recent studies have highlighted the effectiveness of ontology-based decision support frameworks for real-time environmental event detection and intelligent management systems [18][19].

Despite these advances, existing ontology-based air quality frameworks generally do not incorporate weighted rule firing mechanisms, which are essential for capturing the varying influence of pollutants in AQI evaluation.

To the best of our knowledge, no existing outdoor air quality assessment framework that integrates Interval Type-2 fuzzy inference, ontology-based semantic modeling, and toxicity-aware weighted rule firing. The proposed integration simultaneously addresses uncertainty, semantic interpretability, and pollutant severity prioritization, leading to more accurate and regulation-compliant AQI assessment.

## 3. Study Area

The current research work is concentrated on the outdoor air quality index analysis using the monitoring data obtained from the CPCB, India. The area of study includes urban and semi-urban areas that fall under the National Air Quality Monitoring Programme (NAMP), where the ambient air quality index is continuously measured to assess the level of pollution and corresponding health hazards.

The CPCB air quality monitoring stations are set up in a planned manner in cities to identify the spatial and temporal variations of the major air pollutants. These stations are generally established in residential, commercial, traffic, and industrial areas to account for the varied emission sources and health hazards due to air pollution. The monitoring stations are maintained in a standardized manner to ensure accuracy and reliability of the air pollutant measurements.

The air quality indices measured in the current study include particulate matter $PM_{2.5}$, $PM_{10}$, $NO_2$, $SO_2$, $O_3$, CO, and $NH_3$, which form the basis of the IND-AQI. These air pollutants are chosen due to their potential health hazards to humans and their importance as per the CPCB guidelines.

Urban areas in the study region are known to have high population density, rising traffic, industrial activities, and variations in meteorological conditions throughout the year. These conditions lead to uncertainties in AQI levels, especially when close to the threshold values of the categories. The study region is thus apt for assessing the efficacy of uncertainty-aware air quality assessment models.

The chosen study region is representative of a real-world setting where the efficacy of the proposed weighted interval Type-2 fuzzy ontology-based model can be ascertained. The study region is able to capture the imprecision and uncertainties that exist in real-world air quality conditions by using CPCB-supervised outdoor air quality.

## 4. Proposed Framework and Methodology

The proposed framework is based on a sequential hybrid approach that combines data preprocessing, weighted interval Type-2 fuzzy inference, and ontology-based semantic representation for outdoor air quality assessment. The framework is inspired by the weighted interval Type-2 fuzzy AQI models and Type-2 fuzzy ontology frameworks, but is adapted for CPCB-based outdoor air quality data.

The overall workflow consists of five major stages:

- data preprocessing,

- interval Type-2 fuzzy fuzzification,

- fuzzy rule generation,

- weighted rule firing and AQI computation, and

- ontology construction and semantic reasoning.

The proposed methodology is described in the following subsections. Fig. 1 illustrates the architecture of the proposed integrated model used for air quality assessment.

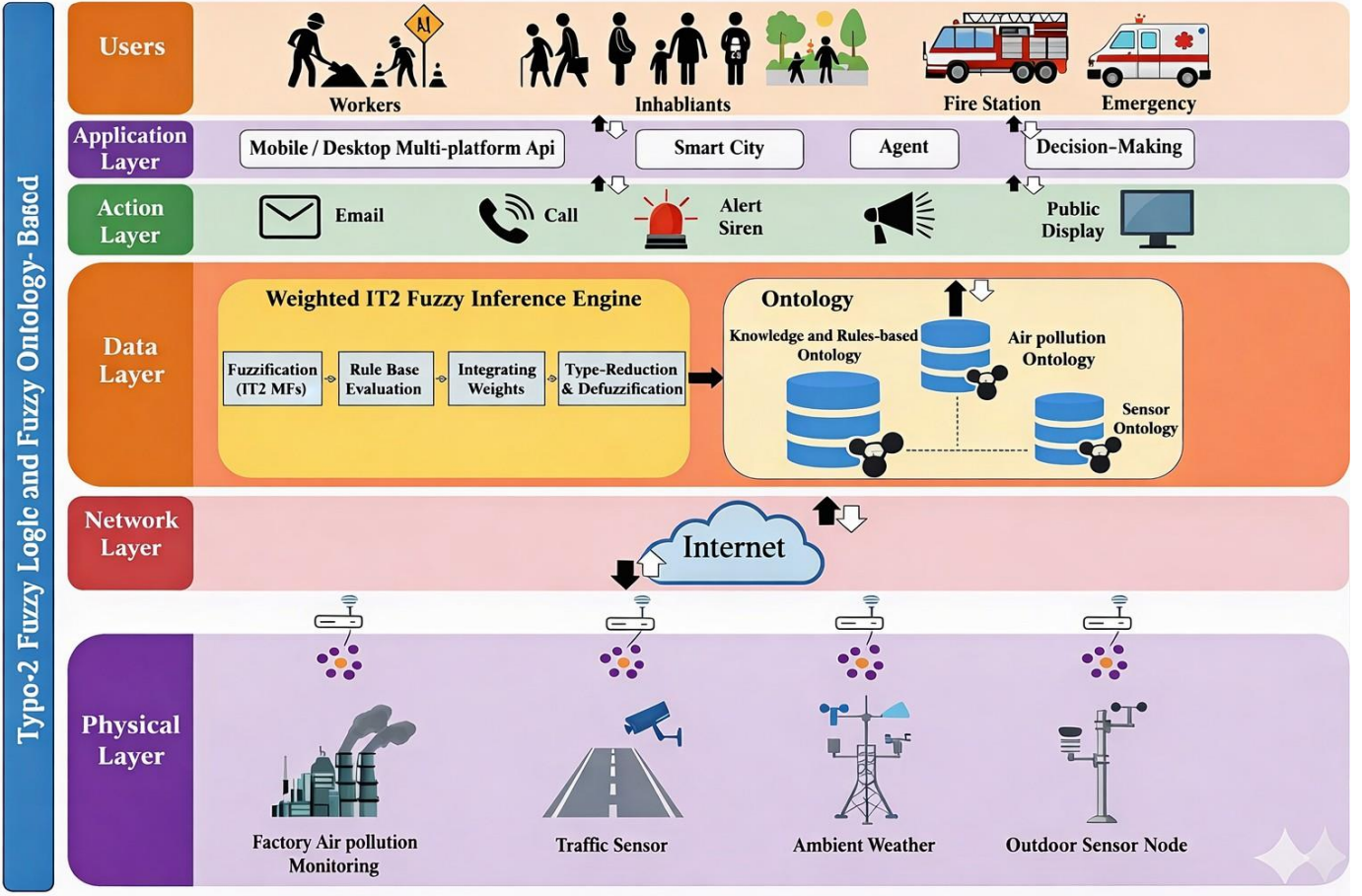

Figure 1: Architecture of the proposed integrated framework for outdoor air quality assessment.

## 4.1. Data Preprocessing

Raw air quality data[1] were collected from CPCB monitoring stations and include multiple pollutant concentration attributes along with AQI labels. Since real-world environmental datasets often contain missing and inconsistent values, a preprocessing stage was applied prior to fuzzy inference.

Fig. 2(a) shows the dataset before preprocessing, highlighting significant missing values across several attributes, which may negatively affect inference accuracy and numerical stability.

To address these data quality issues and ensure reliable AQI classification integrity, records with missing target labels were removed, resulting in the elimination of 21,010 instances. After this step, 87,025 valid instances remained for further analysis. Median-based imputation was then applied to partially missing numerical attributes due to its robustness against outliers and its ability to preserve distributional characteristics. The data were further validated for unit consistency and normalized according to CPCB AQI breakpoint standards to align with fuzzy membership computation.

As illustrated in Fig. 2(b), the resulting dataset contains complete and consistent attribute values. Overall, the preprocessing phase enhances numerical stability, reduces uncertainty from incomplete data, and prepares a reliable foundation for fuzzification, weighted inference, and ontology-based reasoning.

## 4.2. Interval Type-2 Fuzzy Fuzzification

Uncertainty is inherent in air quality assessment, especially when the concentration of air pollutants is close to the boundaries of AQI classes. This is due to the inaccuracies of sensors, variability in time, spatial variability, and vagueness in the interpretation of the regulatory threshold level. To model these uncertainties, interval Type-2 fuzzy sets (IT2FSs) are used in the proposed approach.

Let $X$ denote the universe of discourse for a pollutant concentration. An interval Type-2 fuzzy set $\tilde{A}$ is defined as:

$$\tilde{A} = \left\{ (x, u), \mu_{\tilde{A}}(x, u) \mid x \in X, u \in J_x \subseteq [0, 1] \right\}, \quad (1)$$

where $J_x$ represents the *footprint of uncertainty* (FOU), bounded by the upper and lower membership functions.

### 4.2.1. Trapezoidal Interval Type-2 Membership Functions

For each pollutant and AQI linguistic category (*Good*, *Satisfactory*, *Moderate*, *Poor*, *Very Poor*, and *Severe*), trapezoidal

---

[1] https://www.kaggle.com/datasets/rohanrao/air-quality-data-in-india

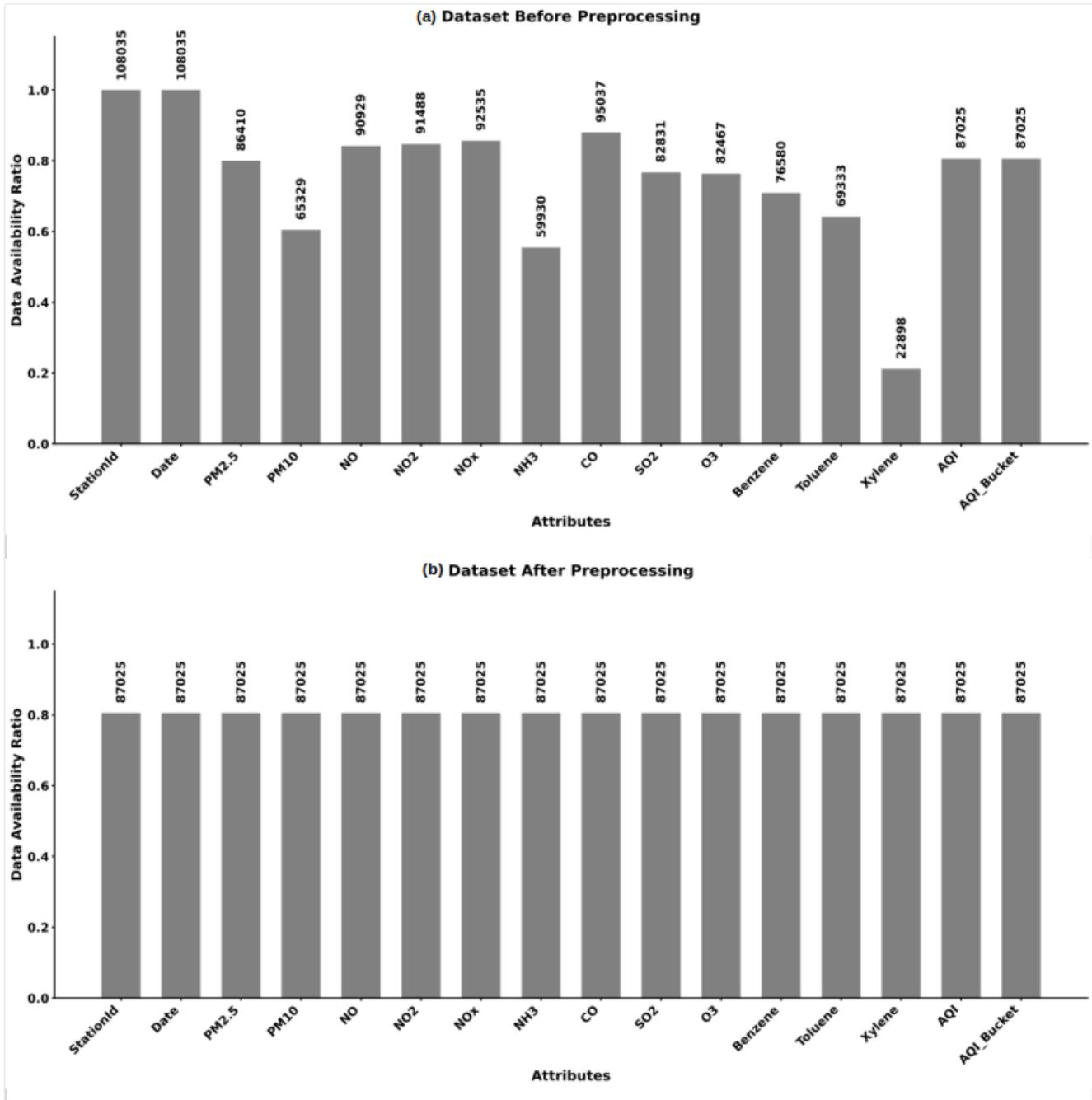

Figure 2: Comparison of data availability ratios before and after preprocessing: (a) dataset before preprocessing, showing missing pollutant values; (b) dataset after preprocessing, demonstrating complete attribute availability.

interval Type-2 membership functions are adopted. Trapezoidal functions are chosen due to their computational simplicity, interpretability, and suitability for representing regulatory breakpoint ranges.

The upper membership function (UMF) of an IT2 fuzzy set $\tilde{A}$ is defined as:

$$\mu_{\tilde{A}}^{U}(x) = \begin{cases} 0, & x < a_u, \\ \frac{x - a_u}{b_u - a_u}, & a_u \leq x < b_u, \\ 1, & b_u \leq x < c_u, \\ \frac{d_u - x}{d_u - c_u}, & c_u \leq x < d_u, \\ 0, & x \geq d_u, \end{cases} \quad (2)$$

where $(a_u, b_u, c_u, d_u)$ are the trapezoidal parameters defining the UMF.

Similarly, the lower membership function (LMF) is defined using parameters $(a_l, b_l, c_l, d_l)$ as:

$$\mu_{\tilde{A}}^{L}(x) = \begin{cases} 0, & x < a_l, \\ \frac{x - a_l}{b_l - a_l}, & a_l \leq x < b_l, \\ 1, & b_l \leq x < c_l, \\ \frac{d_l - x}{d_l - c_l}, & c_l \leq x < d_l, \\ 0, & x \geq d_l. \end{cases} \quad (3)$$

Fig. 3 illustrates the linguistic AQI categories modeled using interval Type-2 fuzzy sets, where overlaps between adjacent linguistic terms (Good, Satisfactory, Moderate, Poor, Very Poor, Severe) represent uncertainty near regulatory thresholds. Fig.4 depicts the structure of a trapezoidal IT2 membership function, highlighting the footprint of uncertainty formed by the UMF and lower membership functions (LMF) bounded by parameters $(a_u, b_u, c_u, d_u)$ and $(a_l, b_l, c_l, d_l)$..

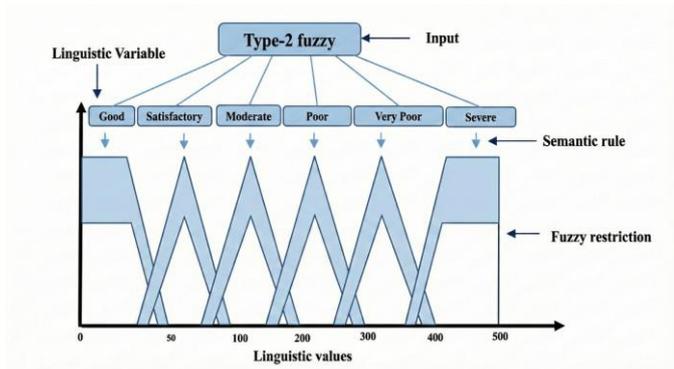

Figure 3: Linguistic representation of AQI categories using interval Type-2 fuzzy sets.

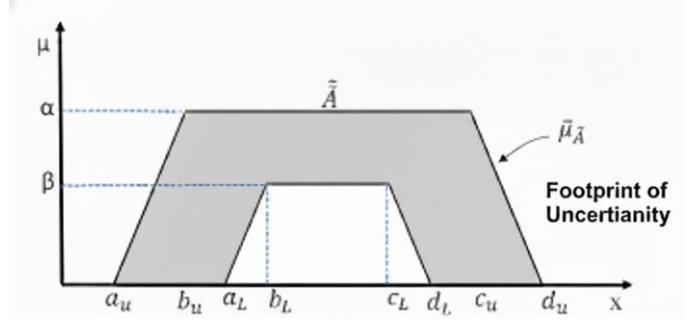

Figure 4: Trapezoidal interval Type-2 membership function

The region enclosed between the UMF and LMF constitutes the footprint of uncertainty (FOU), which explicitly captures the imprecision associated with pollutant concentration boundaries.

The triangular membership function is considered a special case of the trapezoidal membership function when $b = c$. This representation allows for flexibility in modeling both sharp and gradual transitions between AQI classes.

The trapezoidal interval Type-2 membership function parameters used in the fuzzification of air pollutants are presented in Table 1. These parameters are determined based on CPCB AQI breakpoint values, with uncertainty margins accounted for through the lower membership functions to represent real-world variability.

During fuzzification, a crisp pollutant concentration value is transformed into an interval-valued membership degree for each linguistic AQI class by evaluating both the UMF and LMF. These interval membership values are subsequently used in fuzzy rule evaluation and weighted inference, as described in the following subsections.

### 4.3. Interval Type-2 Fuzzy Rule Base Generation

The rule base constitutes the core knowledge component of the proposed interval Type-2 fuzzy inference system (IT2-FIS). Expert knowledge describing the relationship between ambient pollutant concentrations and overall air quality is encoded as fuzzy IF–THEN rules. These rules enable the inference engine to derive the AQI under uncertain and imprecise environmental conditions. The inferred AQI values are subsequently stored in the ontology layer to support semantic interpretation and decision-making.

Let the system consist of $p$ input variables and one output variable. Assuming that $M$ fuzzy rules are defined in the knowledge base, the $i^{th}$ rule is expressed as:

$$R_i : \text{IF } x_1 \text{ is } \tilde{F}_1^i \text{ AND } \ldots \text{ AND } x_p \text{ is } \tilde{F}_p^i, \text{ THEN } y \text{ is } \tilde{G}^i, \quad (4)$$

where $x_j$ denotes the $j^{th}$ input variable, $y$ represents the AQI output, $\tilde{F}_j^i$ and $\tilde{G}^i$ are the interval Type-2 fuzzy sets associated with the antecedents and consequent of rule $i$, respectively.

Defining all possible rule combinations leads to excessive computational complexity, as the number of rules grows exponentially with the number of inputs. Therefore, an *optimal*

Table 1: Membership function parameters [11]

| Pollutant | Good | Satisfactory | Moderate | Poor | Very Poor | Severe |
|---|---|---|---|---|---|---|
| $PM_{2.5}$ | [0,0,15,30;1] [0,0,12,27;0.8] | [15,30,30,60;1] [18,30,30,57;0.8] | [30,60,60,90;1] [33,60,60,87;0.8] | [60,90,90,120;1] [63,90,90,117;0.8] | [90,120,120,250;1] [93,120,120,247;0.8] | [120,250,500,500;1] [123,253,500,500;0.8] |
| $PM_{10}$ | [0,0,25,75;1] [0,0,21,71;0.8] | [25,75,75,175;1] [29,75,75,171;0.8] | [75,175,175,300;1] [79,175,175,296;0.8] | [175,300,300,390;1] [179,300,300,386;0.8] | [300,390,390,470;1] [304,390,390,466;0.8] | [390,470,500,500;1] [394,474,500,500;0.8] |
| $NO_2$ | [0,0,20,60;1] [0,0,16,56;0.8] | [20,60,60,130;1] [24,60,60,126;0.8] | [60,130,130,230;1] [64,130,130,226;0.8] | [130,230,230,340;1] [134,230,230,336;0.8] | [230,340,340,460;1] [234,340,340,456;0.8] | [340,460,500,500;1] [344,464,500,500;0.8] |
| $O_3$ | [0,0,25,75;1] [0,0,21,71;0.8] | [25,75,75,134;1] [29,75,75,130;0.8] | [75,134,134,188;1] [79,134,134,184;0.8] | [134,188,188,478;1] [138,188,188,474;0.8] | [188,478,478,1018;1] [192,478,478,1014;0.8] | [478,1018,1200,1200;1] [482,1022,1200,1200;0.8] |
| $CO$ | [0,0,0.5,1.5;1] [0,0,1,1.1;0.8] | [0.5,1.5,1.5,6;1] [0.9,1.5,1.5,6;0.8] | [1.5,6,6,13.5;1] [1.9,6,6,13.1;0.8] | [6,13.5,13.5,25.5;1] [6.4,13.5,13.5,25.1;0.8] | [13.5,25.5,25.5,42.5;1] [13.9,25.5,25.5,42.1;0.8] | [25.5,42.5,50,50;1] [25.9,42.9,50,50;0.8] |
| $SO_2$ | [0,0,20,60;1] [0,0,16,56;0.8] | [20,60,60,230;1] [24,60,60,226;0.8] | [60,230,230,690;1] [64,230,230,596;0.8] | [230,690,690,1200;1] [234,690,690,1196;0.8] | [690,1200,1200,2000;1] [694,1200,1200,1996;0.8] | [1200,2000,2400,2400;1] [1204,2004,2400,2400;0.8] |
| $NH_3$ | [0,0,200,400;1] [0,0.1,180,360;0.8] | [200,400,400,800;1] [220,400,400,760;0.8] | [400,800,800,1200;1] [420,800,800,1160;0.8] | [800,1200,1200,1800;1] [820,1200,1200,1760;0.8] | [1200,1800,1800,2400;1] [1220,1800,1800,2360;0.8] | [1800,2400,3000,3000;1] [1820,2420,3000,3000;0.8] |
| AQI | [0,0,50,100;1] [0,0,45,95;0.8] | [50,100,100,200;1] [55,100,100,195;0.8] | [100,200,200,300;1] [105,200,200,295;0.8] | [200,300,300,400;1] [205,300,300,395;0.8] | [300,400,400,500;1] [305,400,400,495;0.8] | [400,500,500,600;1] [405,505,500,600;0.8] |

*rule selection strategy* is adopted. For each pollutant, the minimum and maximum observed concentrations during the study period are first identified, and only the relevant AQI linguistic categories are retained. All feasible combinations of these recognized linguistic descriptors are then used to generate the fuzzy rules, ensuring realistic coverage of air quality scenarios while avoiding redundant rules. Representative examples of the generated rules are shown in Table 2.

Table 2: Rule base of the proposed model.

| Rule no. | Rules |
|---|---|
| Rule 1 | IF $PM_{2.5}$ is *Poor* and $PM_{10}$ is *Poor* and $NO_2$ is *Moderate* and $SO_2$ is *Good* and $O_3$ is *Good* and CO is *Good* and $NH_3$ is *Good* then AQI is *Poor*. |
| Rule 2 | IF $PM_{2.5}$ is *Good* and $PM_{10}$ is *Poor* and $NO_2$ is *Satisfactory* and $SO_2$ is *Good* and $O_3$ is *Satisfactory* and CO is *Good* and $NH_3$ is *Good* then AQI is *Poor*. |
| Rule 3 | IF $PM_{2.5}$ is *Very Poor* and $PM_{10}$ is *Poor* and $NO_2$ is *Moderate* and $SO_2$ is *Good* and $O_3$ is *Good* and CO is *Satisfactory* and $NH_3$ is *Good* then AQI is *Very Poor*. |
| Rule 4 | IF $PM_{2.5}$ is *Severe* and $PM_{10}$ is *Good* and $NO_2$ is *Good* and $SO_2$ is *Good* and $O_3$ is *Satisfactory* and CO is *Good* and $NH_3$ is *Good* then AQI is *Severe*. |
| Rule 5 | IF $PM_{2.5}$ is *Good* and $PM_{10}$ is *Satisfactory* and $NO_2$ is *Good* and $SO_2$ is *Good* and $O_3$ is *Satisfactory* and CO is *Satisfactory* and $NH_3$ is *Good* then AQI is *Satisfactory*. |

Both antecedents and consequents are modeled using interval Type-2 fuzzy sets, allowing uncertainty arising from sensor variability and linguistic interpretation to be explicitly represented. The generated rule base forms the foundation of the weighted IT2 fuzzy inference mechanism, after which the weighted inference results are semantically represented within the ontology layer.

*4.4. Weight Evaluation and AQI Assessment Using IT2-FAHP and Weighted IT2 Fuzzy Reasoning*

Thus, a set of inference rules has been identified as sufficient to achieve stable and reliable AQI estimation.

This phase of the proposed methodology focuses on assigning relative importance weights to air pollutants using a priority analysis based on the IT2-FAHP. In conventional interval Type-2 fuzzy inference systems, all input parameters contribute equally during rule evaluation through t-norm and t-conorm operations. However, in real-world air quality assessment, pollutants exhibit unequal impacts on human health and the environment. These impacts vary spatially due to differences in topography, emission sources, meteorological conditions, and urban activities. Therefore, assigning appropriate importance weights to air pollutants is essential for realistic and health-relevant AQI evaluation.

Compared to classical AHP and Type-1 FAHP, IT2-FAHP is more effective in capturing both intra-personal and inter-personal uncertainties by representing expert judgments using interval Type-2 fuzzy numbers [20]. This makes IT2-FAHP particularly suitable for environmental decision-making problems characterized by uncertainty and subjectivity.

### 4.4.1. Health-Impact-Based Pollutant Dominance

Based on epidemiological evidence and regulatory guidelines, particulate matter—especially $PM_{2.5}$—has been identified as the most harmful pollutant due to its strong association with cardiovascular and respiratory diseases. Other pollutants such as $PM_{10}$, $CO$, $O_3$, $NO_2$, $SO_2$, and $NH_3$ exhibit varying degrees of health and environmental impact. Accordingly, the dominance order of pollutants considered in this study is:

$$PM_{2.5} > PM_{10} > CO > O_3 > NO_2 > SO_2 > NH_3$$

### 4.4.2. Linguistic Importance Scale

Expert judgments are expressed using a linguistic importance scale, where each term is represented by a trapezoidal interval Type-2 fuzzy number.

### 4.4.3. Interval Type-2 Fuzzy Pair-Wise Comparison Matrix

Based on expert judgment and historical air quality data, the interval Type-2 fuzzy pair-wise comparison matrix is constructed as:

$$\tilde{C} = \tilde{c}_{ij} = \begin{bmatrix} \tilde{c}_{11} & \tilde{c}_{12} & \cdots & \tilde{c}_{1n} \\ \tilde{c}_{21} & \tilde{c}_{22} & \cdots & \tilde{c}_{2n} \\ \vdots & \vdots & \ddots & \vdots \\ \tilde{c}_{n1} & \tilde{c}_{n2} & \cdots & \tilde{c}_{nn} \end{bmatrix}, \quad j = 1, 2, \ldots, n \quad (5)$$

Each element $\tilde{c}_{ij}$ is defined as:

$$\tilde{c}_{ij} = (a_U, b_U, c_U, d_U; \alpha(b_U), \alpha(c_U)), \\ (a_L, b_L, c_L, d_L; \beta(b_L), \beta(c_L)) \quad (5)$$

The reciprocal judgment is given by:

$$\tilde{c}_{ji} = \left(\frac{1}{d_U}, \frac{1}{c_U}, \frac{1}{b_U}, \frac{1}{a_U}; \alpha(b_U), \alpha(c_U)\right), \\ \left(\frac{1}{d_L}, \frac{1}{c_L}, \frac{1}{b_L}, \frac{1}{a_L}; \beta(b_L), \beta(c_L)\right) \quad (6)$$

### 4.4.4. Consistency Verification

The fuzzy comparison matrix is defuzzified using the trapezoidal IT2 defuzzification method (DTraT) [21]:

$$DTraT = \frac{(d_U - a_U) + \alpha(c_U - b_U) + \alpha(b_U)(c_U - b_U)}{4} \\ + \frac{(d_L - a_L) + \beta(c_L - b_L) + \beta(b_L)(c_L - b_L)}{4} \quad (7)$$

The equivalent crisp matrix $C$ is checked for consistency using:

$$CI = \frac{\lambda_{\max} - n}{n - 1}, \quad CR = \frac{CI}{RI} \quad (8)$$

### 4.4.5. Computation and Normalization of Weights

The fuzzy geometric mean of each row is computed as:

$$\tilde{G}_i = \left[\prod_{j=1}^{n} \tilde{c}_{ij}\right]^{1/n} \quad (9)$$

The fuzzy weights are obtained as:

$$\tilde{w}_i = \tilde{G}_i \otimes \left[\sum_{k=1}^{n} \tilde{G}_k\right]^{-1} \quad (10)$$

Defuzzification and normalization yield the final crisp weights:

$$w_i = \frac{w_i^*}{\sum_{i=1}^{n} w_i^*} \quad (11)$$

### 4.4.6. Evaluation of AQI Using Interval Type-2 Weighted Fuzzy Reasoning

For an input vector

$$\mathbf{x}' = (x_{PM_{2.5}}, x_{PM_{10}}, x_{CO}, x_{O_3}, x_{NO_2}, x_{SO_2}, x_{NH_3}), \quad (12)$$

the evaluation of AQI using interval Type-2 weighted fuzzy reasoning is performed through the following steps.

*Step 1: Membership Interval Computation.* Membership intervals of each pollutant concentration are obtained using trapezoidal interval Type-2 membership functions. For the $n^{th}$ linguistic term, the membership interval is expressed as:

$$\mu_{\tilde{A}_n}(x) = [\underline{\mu}_{\tilde{A}_n}(x), \overline{\mu}_{\tilde{A}_n}(x)], \quad (13)$$

where $\tilde{A}_n \in \{$Good, Satisfactory, Moderate, Poor, Very Poor, Severe$\}$.

*Step 2: Rule Firing Interval.* The firing interval of the $n^{th}$ fuzzy rule is computed using the minimum $t$-norm across all input membership intervals:

$$F_n(\mathbf{x}') = \min\left(\underline{\mu}_{\tilde{A}_{PM_{2.5}}}(x_{PM_{2.5}}), \underline{\mu}_{\tilde{A}_{PM_{10}}}(x_{PM_{10}}), \\ \underline{\mu}_{\tilde{A}_{CO}}(x_{CO}), \underline{\mu}_{\tilde{A}_{O_3}}(x_{O_3}), \underline{\mu}_{\tilde{A}_{NO_2}}(x_{NO_2}), \\ \underline{\mu}_{\tilde{A}_{SO_2}}(x_{SO_2}), \underline{\mu}_{\tilde{A}_{NH_3}}(x_{NH_3})\right), \\ \min\left(\overline{\mu}_{\tilde{A}_{PM_{2.5}}}(x_{PM_{2.5}}), \overline{\mu}_{\tilde{A}_{PM_{10}}}(x_{PM_{10}}), \\ \overline{\mu}_{\tilde{A}_{CO}}(x_{CO}), \overline{\mu}_{\tilde{A}_{O_3}}(x_{O_3}), \overline{\mu}_{\tilde{A}_{NO_2}}(x_{NO_2}), \\ \overline{\mu}_{\tilde{A}_{SO_2}}(x_{SO_2}), \overline{\mu}_{\tilde{A}_{NH_3}}(x_{NH_3})\right) \\ = [f_n^l, f_n^r]. \quad (14)$$

*Step 3: Weighted Rule Firing Interval.* The weighted firing interval of the $n^{th}$ rule is obtained by multiplying the firing interval with the corresponding pollutant weight:

$$F_n^w(\mathbf{x}') = w_n \cdot [f_n^l, f_n^r], \quad (15)$$

where $w_n \in \{w_{PM_{2.5}}, w_{PM_{10}}, w_{CO}, w_{O_3}, w_{NO_2}, w_{SO_2}, w_{NH_3}\}$. If multiple pollutants influence the rule output, the weight with the highest priority is selected to ensure dominant health impact representation.

Table 3: Evidence-based health impact ranking of major air pollutants with permissible exposure limits (WHO/CPCB guidelines)

| Pollutant | Dominant Health Impact | Permissible Limit (WHO/CPCB) |
|---|---|---|
| $PM_{2.5}$ | Fine particulate matter capable of penetrating deep into pulmonary alveoli and bloodstream. Long-term exposure is strongly linked with cardiovascular diseases, stroke, lung cancer, and chronic respiratory disorders. According to WHO guidelines, $PM_{2.5}$ represents the highest mortality and morbidity burden among air pollutants. | 15 $\mu g/m^3$ (24-hour, WHO 2021) |
| $PM_{10}$ | Coarse particulate matter that deposits primarily in the upper respiratory tract. Associated with bronchitis, reduced lung function, aggravated asthma, and cardiovascular complications. WHO and CPCB recognize $PM_{10}$ as a major contributor to respiratory morbidity. | 45 $\mu g/m^3$ (24-hour, WHO 2021) |
| CO | A colorless and odorless toxic gas that combines with hemoglobin. It causes acute toxicity, hypoxia, dizziness, and neurological damage, which can be fatal when inhaled in high concentrations. The World Health Organization has set standards for exposure limits to avoid cardiovascular stress and oxygen deficiency. | 4 $mg/m^3$ (24-hour, WHO 2021) |
| $O_3$ | A secondary photochemical air pollutant resulting from nitrogen oxide and volatile organic compounds. Induces oxidative damage, airway inflammation, lung injury, and aggravates asthma and chronic obstructive pulmonary disease (COPD). WHO recognizes ground-level ozone as a priority urban respiratory risk. | 100 $\mu g/m^3$ (8-hour, WHO 2021) |
| $NO_2$ | A highly reactive nitrogen oxide that irritates and inflames the respiratory system. Precursor to secondary air pollutants such as ozone and fine particulate matter. Chronic exposure enhances vulnerability to respiratory infections and asthma in susceptible groups, including children and the elderly. | 25 $\mu g/m^3$ (24-hour, WHO 2021) |
| $SO_2$ | A highly reactive sulfur oxide gas associated with bronchoconstriction, airway inflammation, and respiratory distress. Short-term exposure can trigger asthma attacks and chronic respiratory disease aggravation. CPCB identifies $SO_2$ as a major industrial emission pollutant. | 40 $\mu g/m^3$ (24-hour, WHO 2021) |
| $NH_3$ | Ammonia primarily affects human health indirectly through secondary particulate formation (ammonium sulfate and ammonium nitrate). Contributes to atmospheric aerosol formation, respiratory irritation, and ecosystem acidification. CPCB monitors $NH_3$ as an emerging pollutant in agricultural and urban environments. | 400 $\mu g/m^3$ (24-hour, CPCB NAAQS) |

Table 4: Linguistic importance scale and interval Type-2 fuzzy representation

| Linguistic Variables | Trapezoidal Fuzzy Numbers |
|---|---|
| Just Equal (JE) | (1,1,1,1;1),(1,1,1,1;0.8) |
| Weakly Important (WI) | (1,2,3,4;1),(1.4,2.4,2.6,3.6;0.8) |
| Between Weakly and Strongly Important (BWSI) | (2,3,4,5;1),(2.4,3.4,3.6,4.6;0.8) |
| Strongly Important (SI) | (3,4,5,6;1),(3.4,4.4,4.6,5.6;0.8) |
| Between Strongly and Very Strongly Important (BSVI) | (4,5,6,7;1),(4.4,5.4,5.6,6.6;0.8) |
| Very Strongly Important (VSI) | (5,6,7,8;1),(5.4,6.4,6.6,7.6;0.8) |
| Between Very Strongly and Absolutely Important (BVAI) | (6,7,8,9;1),(6.4,7.4,7.6,8.6;0.8) |
| Absolutely Important (AI) | (7,8,9,9;1),(7.4,8.4,8.6,9;0.8) |

*Step 4: Type-Reduction using Karnik–Mendel (KM) Algorithm.* For a Mamdani interval Type-2 fuzzy system, each rule $n$ produces an interval firing strength:

$$f_n = [\underline{f}_n, \overline{f}_n]. \quad (16)$$

Let the centroid of the $n$-th consequent fuzzy set be:

$$C_n = [c_n^l, c_n^r], \quad (17)$$

where $c_n^l$ and $c_n^r$ are the left and right centroids of the interval Type-2 consequent set.

The Karnik–Mendel (KM) iterative algorithm is applied to compute the left and right endpoints of the type-reduced set [22][23].

Left endpoint ($AQI_L$):

$$AQI_L = \frac{\sum_{n=1}^{k} \overline{f}_n c_n^l + \sum_{n=k+1}^{N} \underline{f}_n c_n^l}{\sum_{n=1}^{k} \overline{f}_n + \sum_{n=k+1}^{N} \underline{f}_n}. \quad (18)$$

Right endpoint ($AQI_R$):

$$AQI_R = \frac{\sum_{n=1}^{k} \underline{f}_n c_n^r + \sum_{n=k+1}^{N} \overline{f}_n c_n^r}{\sum_{n=1}^{k} \underline{f}_n + \sum_{n=k+1}^{N} \overline{f}_n}. \quad (19)$$

Here, the switch point $k$ is determined iteratively such that:

$$c_k^l \leq AQI_L \leq c_{k+1}^l, \quad c_k^r \leq AQI_R \leq c_{k+1}^r. \quad (20)$$

The centroids $c_n^l$ and $c_n^r$ are arranged in ascending order before applying the KM procedure.

*Step 5: Final Defuzzification.* Finally, the crisp AQI value is obtained by averaging the left and right endpoints of the type-reduced interval, as commonly adopted in interval Type-2 fuzzy systems [22]:

$$AQI = \frac{AQI_L + AQI_R}{2}. \quad (21)$$

## 5. Ontology Model

### 5.1. Semantic Sensor Network Ontology

The SSN ontology, developed by the World Wide Web Consortium (W3C) [24], provides a standardized framework for modeling sensors, sensing systems, observations, and their interaction with environmental phenomena. It enables structured representation of sensing processes and supports interoperability among heterogeneous sensor networks.

Table 5: Final expert-driven linguistic pair-wise comparison matrix

|  | $PM_{2.5}$ | $PM_{10}$ | CO | $O_3$ | $NO_2$ | $SO_2$ | $NH_3$ |
|---|---|---|---|---|---|---|---|
| $PM_{2.5}$ | JE | WI | BWSI | SI | BSVI | AI | AI |
| $PM_{10}$ | 1/WI | JE | SI | BSVI | VSI | AI | AI |
| CO | 1/BWSI | 1/SI | JE | BWSI | VSI | SI | AI |
| $O_3$ | 1/AI | 1/BSVI | 1/BWSI | JE | SI | BSVI | AI |
| $NO_2$ | 1/AI | 1/VSI | 1/VSI | 1/SI | JE | WI | BWSI |
| $SO_2$ | 1/AI | 1/AI | 1/VSI | 1/BSVI | 1/WI | JE | WI |
| $NH_3$ | 1/AI | 1/AI | 1/VSI | 1/AI | 1/BWSI | 1/WI | JE |

Table 6: Random indices for different ordered reciprocal matrices

| Order | 1 | 2 | 3 | 4 | 5 | 6 | 7 | 8 | 9 | 10 |
|---|---|---|---|---|---|---|---|---|---|---|
| RI | 0 | 0 | 0.58 | 0.90 | 1.12 | 1.24 | 1.32 | 1.41 | 1.45 | 1.49 |

Table 7: Final normalized pollutant weights obtained using IT2-FAHP

| Pollutant | Weight |
|---|---|
| $PM_{2.5}$ | 0.3597 |
| $PM_{10}$ | 0.2894 |
| CO | 0.1505 |
| $O_3$ | 0.1022 |
| $NO_2$ | 0.0441 |
| $SO_2$ | 0.0335 |
| $NH_3$ | 0.0205 |

The SSN ontology introduces core concepts including Sensor, System, Observation, ObservableProperty, FeatureOfInterest, and Stimulus. These concepts describe how sensors detect environmental changes, generate observations, and relate sensing results to specific environmental features [25]. The ontology is designed to be extensible and supports domain-specific expansions while maintaining semantic consistency through alignment with the DOLCE-Ultra Lite (DUL) foundational ontology [26]. Extensions of SSN for domain-specific applications have been explored in several studies [30].

Fig. 5 presents the core structure of the SSN ontology and the relationships among sensing components. The proposed air quality ontology extends these foundational concepts by incorporating pollutant-specific parameters and air quality assessment indicators.

### 5.2. Implementation of Ontology

The proposed outdoor air quality ontology is developed in Protégé using OWL 2 and extends the SSN ontology conceptual foundation [28]. The ontology is rooted at *owl:Thing*, from which all domain-specific classes are hierarchically derived. The overall class hierarchy is illustrated in Fig. 6, while the complete inter-domain semantic structure is presented in Fig. 7.

The ontology organizes sensing infrastructure, pollutant categorization, AQI assessment, regulatory compliance, governance workflow, and execution mechanisms into modular semantic domains. The architecture is structured as follows:

#### 5.2.1. EnvironmentalObservationDomain

The Domain represents the sensing layer of the ontology and includes *Sensor*, *System*, *Platform*, *Deployment*, *Observation*, *Result*, and *FeatureOfInterest*, following SSN principles [24]. Environmental parameters are structured under *ObservableProperty*, comprising *Pollutant*, *AirPollutantCategory*, *AirQualityIndex*, and *Weather*. The *Pollutant* class includes CPCB-regulated pollutants such as particulate matter (2.5 micrometers), particulate matter (10 micrometers), nitrogen dioxide, sulfur dioxide, ozone, carbon monoxide, and ammonia [2]. *AirPollutantCategory* organizes pollutants into *GaseousPollutant* and *ParticulateMatter* with CPCB AQI-based linguistic levels, while *AirQualityIndex* represents the aggregated air quality state. Meteorological variables (Temperature, Humidity, Wind, AtmosphericPressure, Rainfall, SolarRadiation) are modeled under *Weather*. Interval Type-2 fuzzy membership intervals are embedded within AQI categories to enable uncertainty-aware pollutant classification [10][29].

#### 5.2.2. EnvironmentalAssessmentDomain

This domain captures event severity levels, health risks, and environmental indicators. Classes such as *EventSeverityLevel*, *CriticalViolation*, and *ExtremeHealthRisk* enable semantic interpretation of pollutant states beyond numerical thresholds and link pollutant exposure to public health implications.

#### 5.2.3. EnvironmentalRegulationDomain

This domain captures CPCB standards, compliance status, guideline thresholds, and regulatory responsibilities. Classes such as *AirQualityStandard*, *GuidelineThreshold*, and *ComplianceStatus* bridge observations with national regulatory structures, thus enabling policy-conformant reasoning [2].

#### 5.2.4. EnvironmentalGovernanceDomain

This domain deals with the enforcement process, accountability, and monitoring of compliance. Classes such as *RegulatoryResponsibility*, *EnforcementWorkflow*, and *ComplianceMonitoring* facilitate environmental governance intelligence and organized administrative action.

#### 5.2.5. EnvironmentalExecutionDomain

This domain comprises actuator devices, actions, and suggested interventions. Classes such as *ActuatorDevice* and *Ac-*

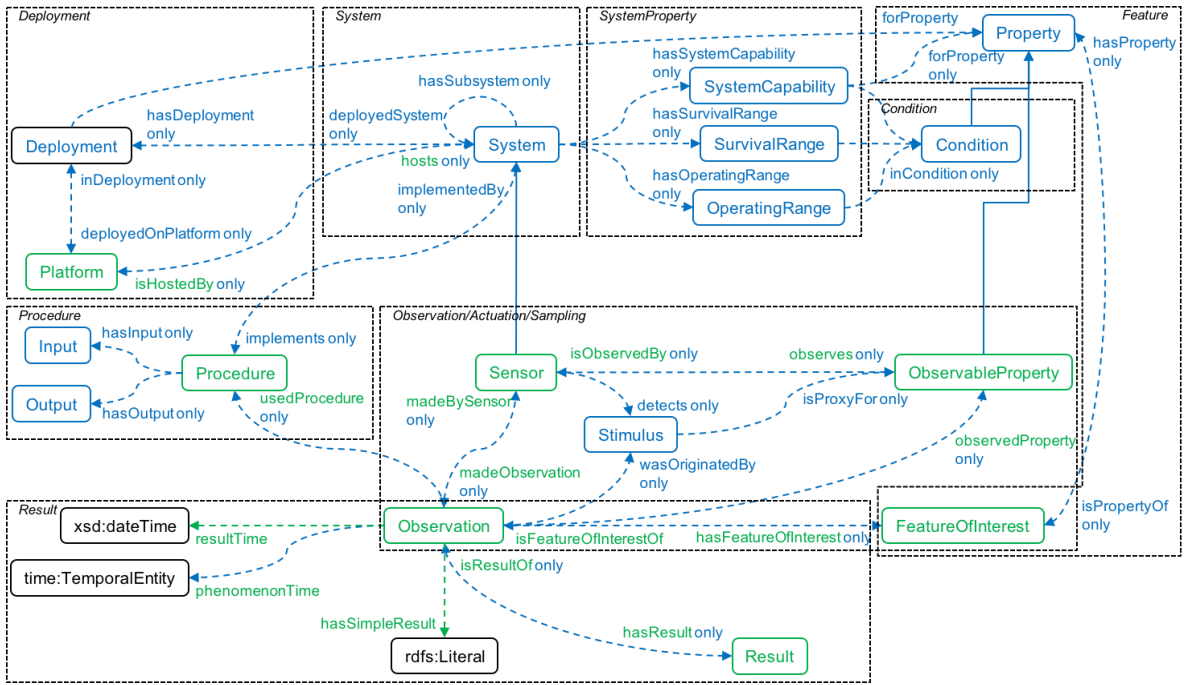

Figure 5: Core architecture of the Semantic Sensor Network ontology

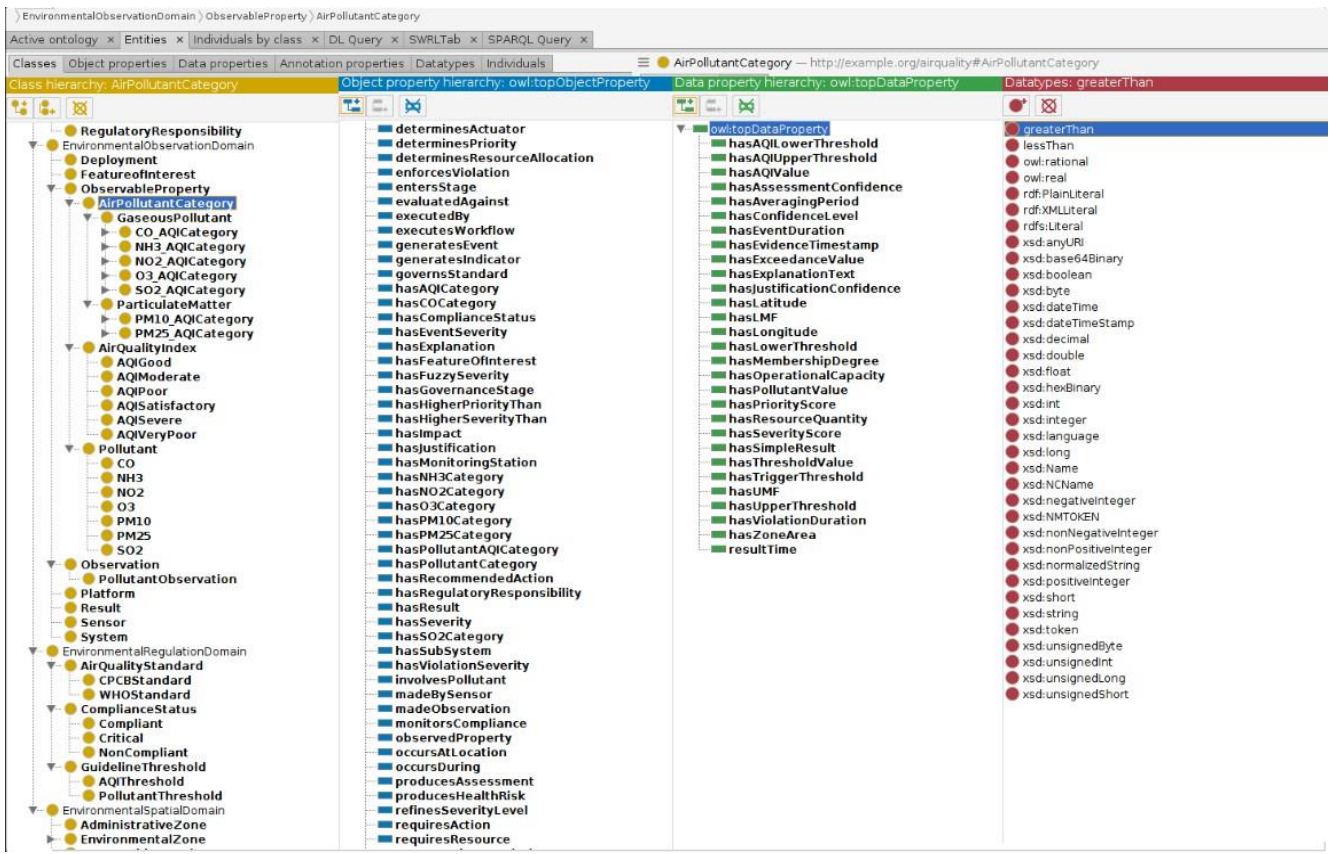

Figure 6: Ontology representation, classes, data and object properties, data types

*tionIndicator* relate inferred AQI states to actionable responses, facilitating automated execution procedures.

### 5.2.6. EnvironmentalExplainabilityDomain

This domain facilitates traceability of rule executions, justification of assessment results, and explanation of decisions via entities such as *RuleTrace* and *AssessmentJustification*. It improves the transparency and explainability of automated semantic reasoning.

The overall ontology graph represented in Fig. 7 illustrates the relationships between classes in observation, assessment, regulation, governance, execution, and explainability domains. This modular and interconnected representation facilitates scalable, regulation-compliant, and semantically robust outdoor air quality assessment.

Supporting sensing and monitoring operations are modeled through Sensor, Platform, System, Observation, Deployment, and Result classes, which collectively represent sensing devices, observation procedures, and measurement outputs. The hierarchical structure ensures modular representation and supports semantic reasoning for air quality monitoring.

### 5.3. SWRL-Based Semantic Reasoning

Semantic reasoning is implemented using the Semantic Web Rule Language (SWRL), enabling logical inference over pollutant categories, AQI states, regulatory implications, and governance actions. The reasoning layer transforms categorized pollutant observations into structured environmental intelligence.

Fig. 8 illustrates a severe pollution observation instance inferred through rule execution.

The primary classification rule is defined as:

```
Observation(?o) ^
hasPM25Category(?o, PM25_Moderate) ^ hasNO2Category(?o, NO2_Good) ^
hasPM10Category(?o, PM10_Severe) ^ hasSO2Category(?o, SO2_Good) ^
hasO3Category(?o, O3_Good) ^ hasCOCategory(?o, CO_Good) ^
hasNH3Category(?o, NH3_Good) -> hasAQICategory(?o, AQISevere)
```

This rule classifies an observation as *AQISevere* when $PM_{10}$ reaches the Severe category, even if other pollutants remain within acceptable limits. The inference reflects regulatory prioritization of high-risk particulate matter exposure.

Once the AQISevere state is inferred, secondary rules propagate health, governance, and execution consequences: As visualized in Fig. 8, the ontology automatically infers:

- Health impacts (premature mortality risk, breathing difficulty),

- Affected vulnerable populations,

- Recommended regulatory actions,

- Responsible enforcement authority,

- Applicable penalties and actuator control.

```
hasAQICategory(?o, AQISevere)
-> hasImpact(?o, PrematureMortalityRisk) ^
   hasImpact(?o, BreathingDifficulty) ^
   affectsVulnerableGroup(?o, AsthmaPatients) ^
   affectsVulnerableGroup(?o, CardiacPatients) ^
   hasRecommendedAction(?o, IndustrialShutdown) ^
   hasRecommendedAction(?o, EmergencyAlertAction) ^
   determinesActuator(?o, IndustrialEmissionController_Instance) ^
   executedBy(?o, StateAuthority) ^
   appliesPenalty(?o, IndustrialClosure)
```

This demonstrates the integration of environmental assessment, governance, and execution domains within a single reasoning workflow.

Additionally, contextual meteorological reasoning is incorporated. For example:

```
WeatherObservation(?w) ^
hasWeatherType(?w, Wind) ^ hasWindSpeed(?w, ?speed) ^
swrlb:lessThan(?speed, 2.0) ^ hasPM10Category(?o, PM10_Severe)
-> hasEscalationEvent(?o, AQIEscalation_Instance)
```

Low wind speed conditions amplify pollutant stagnation, triggering escalation events under severe particulate levels.

Similarly, humidity-based rules can be incorporated to model pollutant dispersion sensitivity under high moisture conditions.

Overall, the SWRL reasoning layer enables automated transformation of pollutant measurements into semantically enriched AQI classification, health risk identification, regulatory response, and governance-driven action execution.

## 6. Results and Discussion

The experimental results show that the proposed ontology-based reasoning framework is more effective than the baseline models in terms of classification reliability, semantic interpretation ability, and adaptability in uncertain situations. The comparison of the proposed framework with the baseline models emphasizes the benefits of combining weighted interval Type-2 fuzzy logic with the ontology-based reasoning process for air quality assessment.

### 6.1. Validation of the Proposed Ontology Model

This section ensures the efficacy of the proposed Outdoor Air Quality ontology in terms of semantic representation, reasoning capability, and knowledge retrieval. The proposed model allows for the semantic representation of air quality observations, inference of AQI categories based on rule-based reasoning, and retrieval of actionable environmental knowledge based on semantic queries. The Pellet reasoner was used to infer implicit knowledge from the proposed ontology, and the validation was carried out by executing SPARQL queries on the inferred knowledge base.

SPARQL Query 1:

```
PREFIX aq: <http://example.org/airquality#>

SELECT ?obs ?aqiValue ?aqiCategory
WHERE {
  ?obs aq:hasStationId aq:CH001 .
  ?obs aq:hasAQIValue ?aqiValue .
  ?obs aq:hasAQICategory ?aqiCategory .
}
```

Figure 7: Comprehensive ontology architecture showing modular domain organization and inter-class relationships

Figure 8: Severe observation instance showing inferred AQI category, health impacts, governance actions, and execution mechanisms

The SPARQL query 1 extracts the AQI value and its inferred category for observations recorded at monitoring station aq:CH001. Fig. 9 presents the RDF graph representation of the inferred knowledge. The successful retrieval of both asserted and inferred properties confirms that the ontology correctly models semantic relationships between observations, monitoring stations, AQI values, and categories.

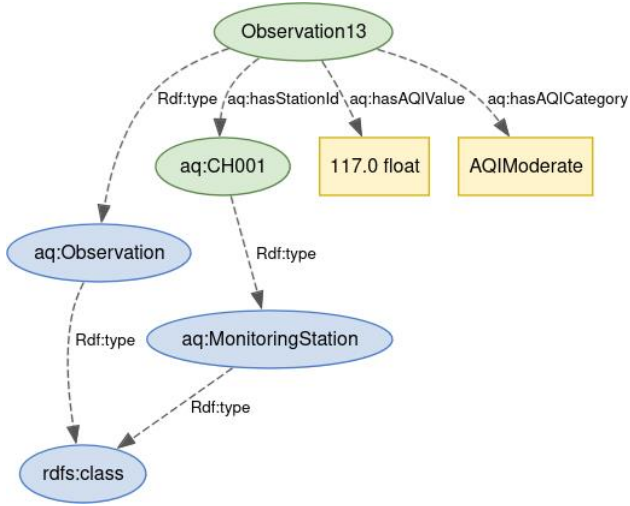

Figure 9: RDF Graph Structure of Inferred Observation

SPARQL Query 2:

```
PREFIX aq: <http://example.org/airquality#>

SELECT ?obs ?station ?aqiValue ?aqiCategory
WHERE {
  ?obs aq:hasRecommendedAction  aq:ConstructionBan .
  ?obs aq:hasStationId ?station .
  ?obs aq:hasAQIValue ?aqiValue .
  ?obs aq:hasAQICategory ?aqiCategory .
}
ORDER BY ?station ?obs
```

The SPARQL query 2 was executed to retrieve all observations associated with the recommended control action aq:ConstructionBan, along with their monitoring station, AQI value, and inferred AQI category. The execution result is illustrated in Fig. 10. This confirms that the ontology successfully infers severe pollution conditions and links them with appropriate mitigation strategies through rule-based reasoning.

DL Query 3:

```
Observation
and hasPM10Category value PM10_Moderate
and hasPM25Category value PM25_Satisfactory
and hasNO2Category value NO2_Good
and hasSO2Category value SO2_Good
and hasO3Category value O3_Good
and hasCOCategory value CO_Good
and hasNH3Category value NH3_Good
and hasStationId value CH001
and hasAQICategory value AQIModerate
and affectsVulnerableGroup value ElderlyVulnerability
and determinesPriority value mediumPriorityInstance
```

A DL Query was executed to validate class membership and inferred relationships for a specific observation instance. The query retrieves an observation recorded at station CH001 with different pollutant category values. It further verifies the inferred AQI category, associated vulnerable group, and the determined response priority. The result, shown in Fig. 11, confirms that the ontology correctly classifies the observation under the AQIModerate category.

Figure 10: SPARQL Query 2 Execution Result Showing Triggered Construction Ban

### 6.2. Ontology Metrics Comparison

To evaluate the structural complexity and knowledge representation capability of the proposed Outdoor Air Quality Ontology, a comparative analysis was conducted against the Indoor Air Quality Domain Ontology developed for the COPD Self-Management System by Ali et al. [31].

Table 8 presents the comparison of core ontology metrics.

Table 8: Ontology Metrics Comparison

| Metric | Proposed Outdoor AQ Ontology | Indoor AQ Ontology[31] |
|---|---|---|
| Total Axioms | 2539 | 407 |
| Logical Axioms | 1320 | 199 |
| Classes | 254 | 34 |
| Object Properties | 78 | 15 |
| Data Properties | 38 | 28 |
| Individuals | 247 | 30 |

The proposed ontology exhibits substantially greater structural richness and domain coverage than the Indoor Air Quality ontology. The higher number of classes (254 vs. 34) reflects broader domain modeling, including pollutants, meteorological factors, AQI categories, monitoring stations, regions, health impacts, and control actions. Similarly, the increased number of object properties (78 vs. 15) and logical axioms (1320 vs.

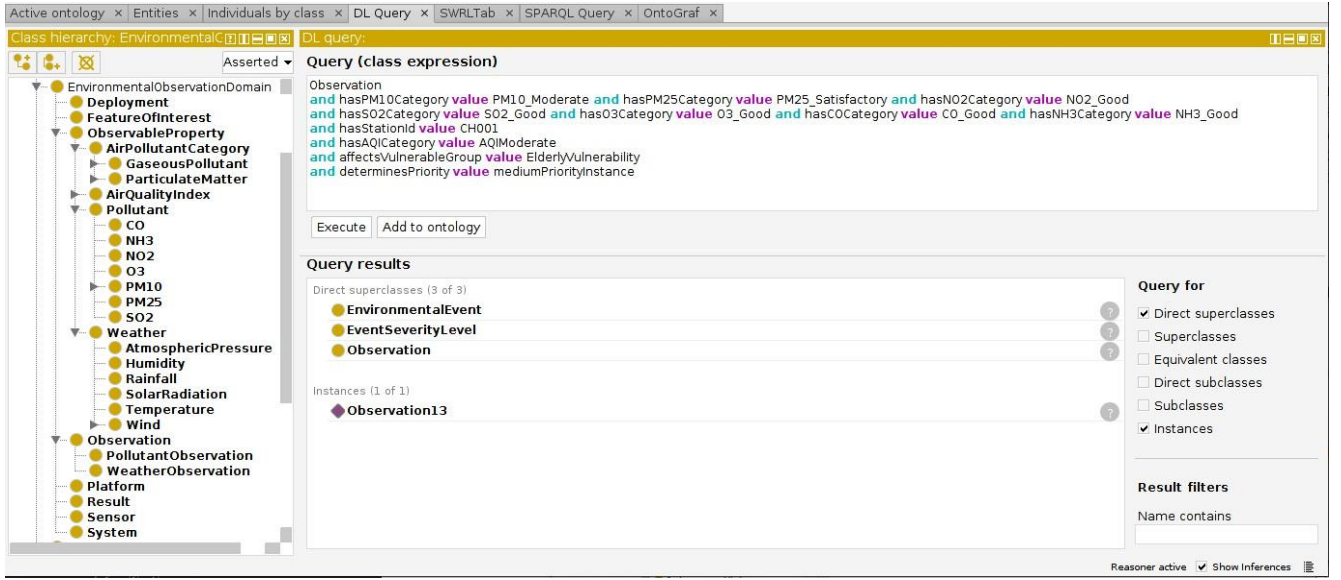

Figure 11: DL Query Result Showing Inferred AQI Category, Vulnerability, and Priority

199) indicates more comprehensive semantic relationships and stronger formal reasoning support. While the Indoor AQ ontology is primarily designed for COPD self-management, the proposed framework addresses large-scale environmental monitoring and semantic decision support, thereby justifying its enhanced structural complexity and expressiveness.

### 6.2.1. Ontology Evaluation based on Score Metrics

Based on the ontology metric scores mentioned in the Table 9, the ontology is evaluated using the quality score measures proposed by [32]. The evaluation is performed using two parameters: the ontology model score ($Score_{om}$), which reflects the structural quality of the ontology, and the knowledge base score ($Score_{kb}$), which measures the efficiency of instance population[33].

Table 9: Ontology Evaluation Scores

| Evaluation Parameter | Score |
|---|---|
| Score om | 24.17 |
| Score kb | 100.97 |

The ontology model score is calculated using Eq. (22):

$$Score_{om} = \frac{(|Rel| \times |Class| \times 100) + ((|Subclass| + |Rel|) \times |Prop|)}{(|Subclass| + |Rel|) \times |Class|} \quad (22)$$

Similarly, the knowledge base score is computed using Eq. (23):

$$Score_{kb} = \frac{(|Class| \times 100) + |Individual|}{|Class|} \quad (23)$$

where $|Individual|$ represents the total number of individuals in the ontology.

Based on the extracted metrics, the computed ontology evaluation scores are presented in Table 9.

### 6.3. Performance Evaluation

Evaluation of the ontology reasoning performance was carried out using widely accepted classification metrics including Precision, Accuracy, Recall, and F1-score. These metrics provide a comprehensive assessment of the reasoning capability and information extraction efficiency of ontology-driven systems. The confusion matrix approach was adopted to quantify system prediction performance, which is commonly utilized for evaluating semantic reasoning and ontology-based retrieval systems [34],[35].

The confusion matrix categorizes the prediction outcomes into four fundamental modes: True Positives (TP), False Positives (FP), True Negatives (TN), and False Negatives (FN). Within the scope of air quality assessment, TP represents cases where unhealthy air conditions are correctly identified by the reasoning framework. TN indicates accurate identification of healthy air conditions. FP corresponds to scenarios where healthy air conditions are incorrectly classified as unhealthy, whereas FN represents instances where unhealthy air conditions are mistakenly categorized as healthy.

The above four prediction outcomes were utilized to compute the evaluation indicators using Eqs. (24)–(27). Precision measures the proportion of correctly predicted unhealthy air quality instances among all predicted unhealthy cases, reflecting the reliability of extracted knowledge. Accuracy evaluates the overall correctness of classification by considering both healthy and unhealthy predictions. Recall quantifies the ability of the system to correctly detect all unhealthy air quality instances present in the dataset. The F1-score combines precision and recall into a single metric, providing a balanced measure of classification performance.

In addition to classification metrics, error-based measures were also employed to evaluate the deviation between predicted and actual AQI categories using Eqs. (28)–(29). The Mean Absolute Error (MAE) measures the average magnitude of predic-tion errors, while the Root Mean Square Error (RMSE) penalizes larger deviations more strongly by squaring the error terms. Lower values of MAE and RMSE indicate better prediction performance.

Table 10: Performance comparison of classical ontology, T2FL fuzzy ontology, and the proposed framework

| Test Samples | Classic Ontology-based System [27] | | | | T2FL indoor Ontology [13] | | | | Proposed Framework | | | | | |
|---|---|---|---|---|---|---|---|---|---|---|---|---|---|---|
| | Pre (%) | Re (%) | Acc (%) | FM (%) | Pre (%) | Re (%) | Acc (%) | FM (%) | Pre (%) | Re (%) | Acc (%) | FM (%) | MAE | RMSE |
| 58 | 54 | 69 | 64 | 60 | 86 | 71 | 73 | 78 | 78 | 77 | 77.23 | 77 | 0.242 | 0.525 |

The mathematical expressions used to compute these metrics are given as follows:

$$Precision = \frac{TP}{TP + FP} \quad (24)$$

$$Accuracy = \frac{TP + TN}{TP + TN + FP + FN} \quad (25)$$

$$Recall = \frac{TP}{TP + FN} \quad (26)$$

$$F1 = \frac{2 \times Precision \times Recall}{Precision + Recall} \quad (27)$$

$$MAE = \frac{1}{n} \sum_{i=1}^{n} |y_i - \hat{y}_i| \quad (28)$$

$$RMSE = \sqrt{\frac{1}{n} \sum_{i=1}^{n} (y_i - \hat{y}_i)^2} \quad (29)$$

where $y_i$ represents the actual AQI category, $\hat{y}_i$ denotes the predicted AQI category, and $n$ is the total number of observations.

The classical ontology-based approach shows moderate classification performance due to its reliance on crisp decision boundaries, whereas the incorporation of Type-2 fuzzy ontology improves uncertainty modeling and overall classification effectiveness. As shown in Table 10, Type-2 fuzzy ontology enhances precision from 54% to 78%, accuracy from 64% to 77%, and F1-score from 60% to 78%. In addition, the proposed model achieves a MAE of 0.242 and a RMSE of 0.525, indicating that the predicted AQI categories deviate by less than one class level on average from the ground truth CPCB AQI categories. These results demonstrate the advantage of fuzzy-based semantic reasoning. The proposed framework further improves classification stability and achieves balanced performance across all evaluation metrics by integrating weighted interval Type-2 fuzzy reasoning, which effectively handles uncertainty near pollutant threshold boundaries and reduces misclassification errors.

*6.4. Generalization and Applicability*

The proposed weighted interval Type-2 fuzzy ontology-based framework is not limited to CPCB AQI standards and can be generalized to other global air quality assessment systems. The modeling approach can be adapted to alternative AQI standards such as the World Health Organization (WHO) guidelines and the United States Environmental Protection Agency (US EPA) AQI by modifying the membership function parameters and regulatory thresholds.

Furthermore, the integration of semantic ontology and uncertainty-aware fuzzy reasoning enables the framework to be deployed in smart city environments, real-time monitoring systems, and IoT-based environmental decision support systems. The modular architecture ensures that the proposed system can be extended to different geographical regions and heterogeneous environmental datasets while maintaining interpretability and decision-making capability.

## 7. Conclusion and Future Work

This study presented a weighted Interval Type-2 fuzzy ontology-based framework for intelligent and uncertainty-aware outdoor air quality assessment. Given that air pollution data are inherently imprecise and exhibit gradual transitions across AQI categories, conventional crisp threshold-based approaches are inadequate for capturing ambiguity near class boundaries. To address this limitation, Interval Type-2 fuzzy sets were employed to explicitly model uncertainty in pollutant classification, while IT2-FAHP-based weighting enabled the incorporation of pollutant-specific health impacts into the inference process. Furthermore, the proposed framework integrates an OWL-based semantic ontology to represent pollutants, monitoring stations, AQI categories, vulnerable populations, and recommended mitigation actions in a structured and machine-interpretable manner. The combination of uncertainty-aware fuzzy reasoning with semantic knowledge modeling enhances interpretability, supports explainable decision-making, and enables intelligent inference through SWRL rules and SPARQL-based validation. Experimental validation using CPCB air quality data confirms that the proposed approach improves AQI classification reliability and effectively captures uncertainty compared to traditional and Type-1 fuzzy methods.

In contrast to existing Type-2 fuzzy ontology-based air quality models, which are primarily focused on indoor air quality assessment, the proposed framework addresses outdoor air quality evaluation under real-world environmental conditions. It introduces health-impact-driven weighted reasoning and CPCB-compliant AQI modeling, thereby providing a more realistic and regulation-aligned assessment of environmental conditions. Moreover, the modular architecture ensures that the framework is generalizable beyond CPCB standards. By adapting membership functions and regulatory thresholds, the model can be extended to global AQI systems such as WHO guidelines and US EPA standards, and can be deployed in smart city and IoT-based environmental monitoring systems. However, the current framework remains semi-automated and may require regional customization due to variations in pollutant behavior and regulatory policies. Additionally, the reliance on predefined expert rules may limit full automation and scalability in highly dynamic environments.

Future work will focus on improving automation and scalability of the framework. Adaptive learning techniques will be introduced to automatically refine rules and parameters based on regional data, reducing manual configuration. Integration with LLMs, AI-driven predictive analytics, Big Data technologies, and knowledge graph expansion will further enhance adaptability, real-time processing, and intelligent environmental decision support.

**CRediT authorship contribution statement**

**Data Availability**

The dataset used in this study, namely the Air Quality Data in India (2015–2020), is publicly available on the Kaggle repository. It contains air quality index (AQI) and pollutant data collected at hourly and daily levels from multiple monitoring stations across various cities in India. The dataset can be accessed at https://www.kaggle.com/datasets/rohanrao/air-quality-data-in-india.

**Declaration of competing interest**

The authors declare that they have no known competing financial interests or personal relationships that could have appeared to influence the work reported in this paper.

**Acknowledgments**

This research was supported by the Council of Science and Technology, Uttar Pradesh (CSTUP), under Sanction No. CST/D-1198. The authors thank CSTUP for the funding and IIIT Allahabad, Prayagraj, for providing infrastructure and institutional support and the Big Data Analytics Lab members for providing valuable suggestions for carrying out this work.

**Appendix**

Table 11: List of Abbreviations

| Abbreviation | Full Form |
|---|---|
| AQI | Air Quality Index |
| IND-AQI | Indian Air Quality Index |
| CPCB | Central Pollution Control Board |
| WHO | World Health Organization |
| IT2FS | Interval Type-2 Fuzzy Set |
| IT2-FIS | Interval Type-2 Fuzzy Inference System |
| IT2-FAHP | Interval Type-2 Fuzzy Analytic Hierarchy Process |
| FAHP | Fuzzy Analytic Hierarchy Process |
| FOU | Footprint of Uncertainty |
| UMF | Upper Membership Function |
| LMF | Lower Membership Function |
| KM | Karnik–Mendel Algorithm |
| DTraT | Defuzzification of Trapezoidal IT2 |
| SSN | Semantic Sensor Network |
| OWL | Web Ontology Language |
| SWRL | Semantic Web Rule Language |
| RDF | Resource Description Framework |
| SPARQL | SPARQL Protocol and RDF Query Language |
| NAMP | National Air Quality Monitoring Programme |
| LLM | Large Language Model |
| PM2.5 | Particulate Matter ($2.5\,\mu$m) |
| PM10 | Particulate Matter ($10\,\mu$m) |
| $NO_2$ | Nitrogen Dioxide |
| $SO_2$ | Sulfur Dioxide |
| $O_3$ | Ozone |
| CO | Carbon Monoxide |
| $NH_3$ | Ammonia |